%% file: main_arxiv.tex
\documentclass[sigconf,authorversion,nonacm,screen]{acmart}

\usepackage[utf8]{inputenc}
\usepackage[T1]{fontenc}

\renewcommand{\sfdefault}{phv}

\normalfont\selectfont

\makeatletter
\renewcommand\operator@font{\sf}
\makeatother

\usepackage{pdfpages}

\usepackage{pifont}

\usepackage{xspace}
\usepackage{epsfig}
\usepackage{adjustbox} %
\graphicspath{{./figures/}}
\DeclareGraphicsExtensions{.pdf,.png,.jpeg,.jpg}

\usepackage{bm} %
\usepackage{esint} %
\usepackage{commath} %

\usepackage{siunitx} %
\usepackage{float} %
\usepackage{stfloats} %

\usepackage{subcaption} %

\usepackage{booktabs} %
\usepackage{array} %

\usepackage{multirow} %
\usepackage{rotating} %

\usepackage{enumitem} %

\usepackage[capitalize]{cleveref} %
\Crefname{section}{Section}{Sections}
\crefname{section}{Sec.}{Secs.}
\Crefname{table}{Table}{Tables}
\crefname{table}{Table}{Tables}

\renewcommand{\mid}{\,\ifnum\currentgrouptype=16 \middle\fi|\,}

\crefname{table}{Table}{Tables}
\Crefname{table}{Table}{Tables}
\usepackage{colortbl} %
\let\oldmarginpar\marginpar
\renewcommand\marginpar[1]{\-\oldmarginpar[\raggedleft\footnotesize #1]%
{\raggedright\footnotesize #1}}

\setcopyright{rightsretained}
\copyrightyear{2023}
\acmYear{2023}
\acmDOI{10.1145/3592458}

\def\Ours{{AutoStory}\xspace}

\begin{document}

\title[Generating Diverse Storytelling Images with Minimal Human Effort]{\rm AutoStory:  Generating Diverse Storytelling Images with Minimal Human Effort}

\author{Wen Wang}
\authornote{Equal Contribution.}
\affiliation{%
 \institution{Zhejiang University}\city{Hangzhou}\country{China}}
\email{wwenxyz@zju.edu.cn}

\author{Canyu Zhao}
\authornotemark[1]
\affiliation{
 \institution{Zhejiang University}\city{Hangzhou}\country{China}}
\email{volcverse@zju.edu.cn}

\author{Hao Chen}
\affiliation{%
 \institution{Zhejiang University}\city{Hangzhou}\country{China}}
\email{haochen.cad@zju.edu.cn}

\author{Zhekai Chen}
\affiliation{
 \institution{Zhejiang University}\city{Hangzhou}\country{China}}
\email{chenzhekai@zju.edu.cn}

\author{Kecheng Zheng}
\affiliation{
\institution{Zhejiang University}
\city{Hangzhou}
\country{China}}
\email{zkechengzk@gmail.com}

\author{Chunhua Shen}
\affiliation{%
 \institution{Zhejiang University}\city{Hangzhou}\country{China}}
\email{chunhuashen@zju.edu.cn}

\begin{abstract}

\it

Story visualization aims to generate a series of images that match the story described in texts, and it requires the generated images to satisfy high quality, alignment with the text description, and consistency in character identities. Given the complexity of story visualization, existing methods drastically simplify the problem by considering only a few specific characters and scenarios, or requiring the users to provide per-image control conditions such as sketches. However, these simplifications render these methods incompetent for real applications. 

To this end, we propose an automated story visualization system that can effectively generate diverse, high-quality, and consistent sets of story images, with minimal human interactions. Specifically, we utilize the comprehension and planning capabilities of large language models for layout planning, and then leverage large-scale text-to-image models to generate sophisticated story images based on the layout. We empirically find that sparse control conditions, such as bounding boxes, are suitable for layout planning, while dense control conditions, {\em e.g.}, sketches and keypoints, are suitable for generating high-quality image content. To obtain the best of both worlds, we devise a dense condition generation module to transform simple bounding box layouts into sketch or keypoint control conditions for final image generation, which not only improves the image quality but also allows easy and intuitive user interactions. 

In addition, we propose a simple yet effective method to generate multi-view consistent character images, eliminating the reliance on human labor to collect or draw character images. This allows our method to obtain consistent story visualization even when only texts are provided as input. Both qualitative and quantitative experiments demonstrate the superiority of our method.

    Project webpage: \url{https://aim-uofa.github.io/AutoStory/}

\end{abstract}



\begin{teaserfigure}
  \centering
  \includegraphics[width=0.985\linewidth]{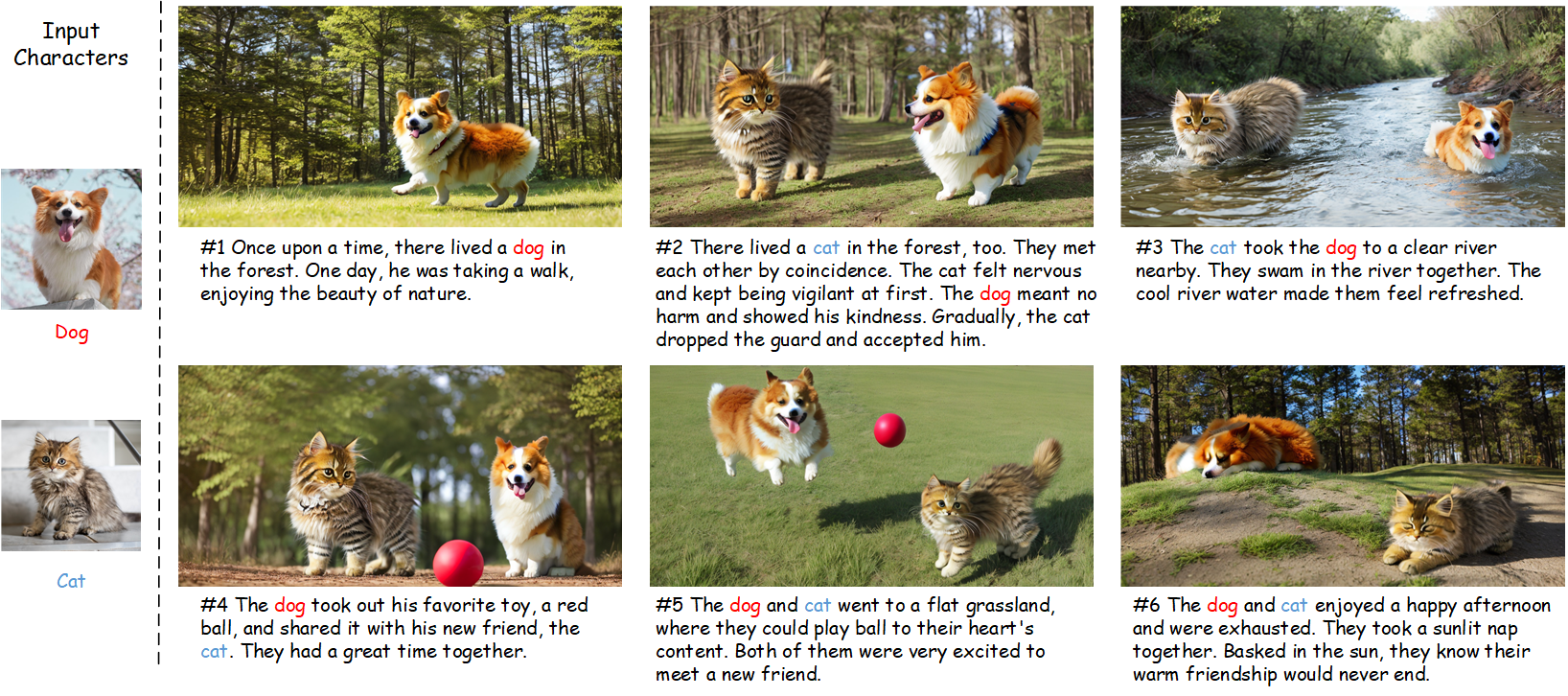}
  \caption{Example storytelling images generated by our method \Ours. We can generate text-aligned, identity-consistent, and high-quality story images from user-input stories and characters (the dog and cat on the left, specified by about 5 images per character), without additional inputs like sketches \cite{gong2023talecrafter}. \textit{
  Further, our method also supports generating storytelling images from only text inputs, as %
  shown
  in our experiments.} 
  }
  \Description{Description}
  \label{fig:teaser}
\end{teaserfigure}

\maketitle %

\section{Introduction}
\label{sec:intro}

Story visualization aims to generate a series of visually consistent images from a story described in text. It has a wide range of applications.
For example, it can provide creativity and inspiration in art creation, and open up new opportunities for artists. In child education, it can stimulate children's imagination and creativity, and make the learning process more interesting and effective. %
In cultural inheritance, it can provide a rich variety of visual expressions for various creative and cultural activities described in texts.

Yet, 
story visualization is a very challenging task, which %
needs to meet 
multiple requirements for the generated images, including (1) high quality: the generated images must be visually appealing and have a reasonable layout (2) consistency: not only the generated images should be consistent to the text descriptions, but also the identities of the characters and scenes in different images should be consistent; and (3) versatility: to satisfy a wide range of users' needs, it needs to be able to be easily applied to different styles, characters, and scenes.

Limited by the capabilities of generative models, previous work    \cite{li2019storygan,maharana2021vlcstorygan,maharana2021ducostorygan,maharana2022storydall} significantly and overly simplifies the 
task 
by considering story visualization for specific styles, scenes, and characters on fixed datasets, such as the PororoSV    \cite{li2019storygan} and FlintstonesSV    \cite{maharana2021vlcstorygan} datasets. Generative models trained on large-scale text-to-image data and few-shot customized generation methods   \cite{ruiz2022dreambooth,textual_inversion} bring new opportunities for story visualization. Some recent work   \cite{gong2023talecrafter,liu2023intelligent} attempts to obtain story visualization for which characters can be generalized, but are still limited to comic book style image production and %
often 
rely on additional user input conditions, such as sketches.

Unlike these efforts, we propose a versatile story visualization method, termed \Ours, that is fully automated and capable of generating high-quality stories with diverse characters, scenes, and styles. Users only need to enter simple story descriptions to generate high-quality storytelling images. 
On the other hand, our method is sufficiently general to accommodate various user inputs, 
providing 
a flexible interface that allows the user to subtly control the outcome of story visualization through simple interactions. For example, depending on the user's needs, the user can control the generated story by providing an image of the character, adjusting the layout of the objects in the picture, adjusting the character's pose, sketching, and so on.

Given the complexity of story scenes, the general idea of our \Ours is to utilize the comprehension and planning capabilities of large language models to achieve layout planning, and then generate complex story scenes based on the layout.
Empirically, we find that sparse control conditions, like bounding boxes, are suitable for layout planning, while dense control conditions, like sketches and keypoints, are suitable for generating high-quality image content. To %
have 
the best of both worlds, we devise a \textit{dense condition generation module} as the bridge. Instead of directly generating the whole complex picture, we first utilize the local prompt generated by the large language model to generate individual subjects in the stories, and then extract the dense control conditions from the subject images. The final story images are generated by conditioning on the dense control signals.
Thus, 
our \Ours effectively utilizes the planning capability of large language models, while ensuring high-quality generation results in a fully automatic
fashion. 
At the same time, we allow users to edit the layout and other control conditions generated by the algorithm to better align with their intentions.

To achieve identity consistency in the generated images while also maintaining the versatile ability of the large-scale text-to-image generative models, unlike existing methods that perform 
time-consuming 
training on domain-specific data, we %
exploit few-shot parameter-efficient fine-tuning techniques for foundation models. Combining with customized generation techniques, 
\Ours achieves identity-consistent generation by training on \textit{only a few images for each character}, while also generalizing to diverse characters, scenes, and styles.

In addition, existing story visualization methods require the user to provide multiple images for each character in the story, 
which 
need to be 
both identity consistent and diverse. This can be laborious since the users have to draw or collect multiple images for each character. We eliminate this requirement by proposing a multi-view consistent subject generation method. Specifically, we propose a training-free identity consistency modeling method by
treating multiple views as a video and jointly generating the textures with temporal-aware attention. Furthermore, we improve the diversity of the generated character images by leveraging the 3D prior in view-conditioned image translation models    \cite{liu2023zero1to3,liu2023one}, without compromising identity consistency.
An %
example 
story visualization 
is shown in Fig.~\ref{fig:teaser}.

To summarize, our main contributions are as follows.
\begin{itemize}
    
    \item 
    We propose a fully automated story visualization pipeline that can generate diverse, high-quality, and consistent stories with minimal user input requirements.
    
    \item
    To 
    deal with 
    complex scenarios in story visualization, we leverage sparse control signals for layout generation, while dense control signals for high-quality image generation. A simple yet effective dense condition generation module is proposed as the bridge to transform sparse control signals into sketch or keypoint control conditions 
    fully automatically. 
    
    \item 
    To maintain identity and eliminate the need for users to draw or collect image data for characters, we propose a simple method to generate multi-view consistent images from only texts. Specifically, we use a 3D-aware generative model to improve the diversity and generate identity-consistent data by viewing the images from multiple views as a video.
    
    \item
    To our knowledge, we develop the first method which is able to generate high-quality storytelling images in diverse characters, scenes, and styles, even when the user inputs only text.
    Simultaneously, our method is flexible to accommodate 
    various user inputs where needed.
\end{itemize}

\section{Related Work}
\label{sec:related_work}

\subsection{Story Visualization}

Story visualization aims to generate a series of visually consistent images from a story described in text. Limited by the generative %
capacity 
of the model, many story visualization approaches    \cite{li2019storygan,song2020cpcsv,li2022word,chen2022character,maharana2021ducostorygan,maharana2021vlcstorygan,maharana2022storydall,pan2022synthesizing,rahman2022make} seek to 
largely 
simplify the task such that it becomes tractable, by considering specific characters, scenes, and image styles in a particular dataset. 
Early story visualization methods are mostly built upon GANs~\cite{goodfellow2020generative}. For example, StoryGAN    \cite{li2019storygan} pioneers the story visualization task by proposing a GAN-based framework that considers both the full story and the current sentence for coherent image generation. CP-CSV    \cite{song2020cpcsv}, DuCo-StoryGAN    \cite{maharana2021ducostorygan}, and VLC-StoryGAN    \cite{li2019storygan} follow the GAN-based framework, while improving the consistency of storytelling via better character-preserving or text understanding. 
Difference from these works, VP-CSV    \cite{chen2022character} leverages VQ-VAE a transformer-based language model for story visualization. StoryDALL-E    \cite{maharana2022storydall} leverages the pre-trained DALL-E    \cite{ramesh2021zero} for better story visualization and proposes a novel task named story continuation that supports story visualization with a given initial image. AR-LDM    \cite{pan2022synthesizing} proposes a diffusion model-based method that generates story images in an auto-regressive manner.

While progress has been made, these methods rely on story-specific training on datasets like PororoSV    \cite{li2019storygan} and FlintstonesSV    \cite{maharana2021vlcstorygan}, making it difficult to generalize these methods to varying characters and scenes.

The development of large-scale pre-trained text-to-image generative models    \cite{ramesh2021zero,ramesh2022hierarchical,rombach2022high,saharia2022photorealistic} opens up new opportunities for generalizable story visualization. 
Several attempts have been made to generate storytelling images with diverse characters    \cite{jeong2023zero,liu2023intelligent,gong2023talecrafter}. %
Jeong et al.\     \cite{jeong2023zero} utilized textual inversion    \cite{textual_inversion} to swap human identities in story images, thus generalizing the characters in story visualization. However, the identity is not well preserved, and the method is limited to a single human character in storytelling.
Intelligent Grimm    \cite{liu2023intelligent} proposes the task of open-ended visual storytelling. They collect a dataset of children’s storybooks and train an autoregressive generative model for story visualization. %
The limitation is clear: 
they focus on the storytelling of the children’s storybook style, and it needs to re-train the model to generalize to other styles, contents, etc., which is not scalable.

Probably the most similar work to ours is TaleCraft    \cite{gong2023talecrafter}, which also proposes a systematic pipeline for story visualization. %
Note that, 
they require user-provided sketches for each character in each story image to obtain visually pleasing generations, which can be laborious to obtain. 
Moreover, all existing methods rely on multiple user-provided images for each character to obtain identity-coherent story visualizations. In contrast, our method allows for generating diverse and coherent story visualization results with only text descriptions as inputs.

\subsection{Controllable Image Generation}
The scaling of text-image paired data    \cite{schuhmann2022laion}, computational resources, and model size have enabled unprecedented text-to-image (T2I) generation results    \cite{ramesh2021zero,ramesh2022hierarchical,saharia2022photorealistic,rombach2022high}.
Large-scale pre-trained text-to-image models, such as Stable Diffusion \cite{rombach2022high}, are capable of generating images from text, {\em i.e.}, $I=
\operatorname{DM} 
\left(p\right)$, where $\operatorname{DM}(\cdot)$ is the pre-trained diffusion model and $p$ is the text prompt that describe the image $I$. 
In this process, the text information is passed into the image's latent representation through cross-attention layers in the model. The attention  \cite{vaswani2017attention} operation can be written as:
\begin{equation}
    \operatorname{Attn}(Q, K, V)=\operatorname{Softmax}\left(\frac{Q K^T}{\sqrt{d}}\right) \cdot V,
\end{equation}
with $Q=W^Q z_{i}, K=W^K \operatorname{Enc}(p), V=W^V \operatorname{Enc}(p)$. Here, $W^Q$, $W^K$, and $W^V$ are the projection weights 
of 
the attention layer, respectively. $\operatorname{Enc}(\cdot)$ is the text encoder, and $z_{i}$ is the latent image feature.

However, limited by the language understanding capability of the text encoder and poor text-to-image content association  \cite{chefer2023attendandexcite}, T2I models, like Stable Diffusion    \cite{rombach2022high}, can perform poorly in the generation of multiple characters and complex scenes    \cite{chefer2023attendandexcite}. To alleviate this drawback, some approaches introduce explicit spatial guidance in T2I generative models. For example, ControlNet    \cite{zhang2023adding} uses zero convolution layers and a trainable copy of the original model weights, introducing reliable control in 
diffusion models.
T2I-Adapter    \cite{mou2023t2i} achieves control ability by proposing the adapter that extracts guidance feature and adds it %
to 
the feature from the corresponding UNet encoder.
GLIGEN    \cite{li2023gligen} injects a gated self-attention block into the UNet, %
enabling
the model to make good use of the grounding inputs.

Inspired by the ability of large language models (LLMs)    \cite{openai2023gpt4,anil2023palm} 
being able to 
understand and plan, recent works    \cite{feng2023layoutgpt,lian2023llmgrounded} %
employ 
LLMs for layout generation. Specifically, LayoutGPT    \cite{feng2023layoutgpt} achieves plausible results in 2D image layouts and even 3D indoor scene synthesis by applying in-context learning on LLMs. 
LLM-grounded Diffusion   \cite{lian2023llmgrounded} proposes a two-stage process based on the LLM-generated layout and local prompts. Specifically, it first generates the local objects within each bounding box based on the corresponding local prompt, and then re-generates the final result based on the inversed latent of local objects. 
While effective, LLM-grounded Diffusion requires careful hyper-parameter tuning for the trade-off between structural guidance and inter-object relationship modeling. Moreover, it is difficult for the users to control the detailed structure of the generated objects. 
In contrast, we use the intuitive sketch or keypoint to guide the final image generation. 
Thus, we can not only achieve high-quality story image generation, but also allow interactive story visualization by simply tuning the generated sketch or keypoint conditions.

\subsection{Customized Image Generation}

Story visualization requires that the identities of characters and scenes in a story remain consistent across different images. Customized image generation can meet this requirement to a large extent. Early methods    \cite{ruiz2022dreambooth,textual_inversion} focus on the customized generation of a single object. For example, DreamBooth    \cite{ruiz2022dreambooth} fine tunes the pre-trained T2I diffusion model under a class-specific prior-preservation loss. Textual Inversion    \cite{textual_inversion} enables customized generation by inverting subject image content into text embeddings. Unlike these approaches, Custom Diffusion    \cite{kumari2022multi} further achieves multi-subject customization by combining the multiple customization weights through closed-form constrained optimization. Cones    \cite{liu2023cones} finds that a small cluster of concept neurons in the diffusion model corresponds to a single subject, and thus achieves customized generation of multiple objects by combining these concept neurons. Cones2    \cite{liu2023cones2} further achieves more effective multi-object customization by combining text embedding of different concepts with simple layout control. Differently, Mix-of-Show    \cite{gu2023mix} proposes gradient fusion to effectively combine multiple customized LoRA    \cite{hu2022lora} weights and performs multi-object customization with the aid of the T2I-Adapter's dense controls.

While significant progress has been made, existing methods perform poorly on \textit{one-shot} customization. The training data for subject-driven generation has to be identity-consistent and diverse. As a result, existing story visualization methods require multiple user-provided images for each character. To tackle this issue, we propose \textit{a training-free consistency modeling method}, and leverage the 3D prior in 3D-aware generative models    \cite{liu2023zero1to3,liu2023one} to obtain multi-view consistent character images for customized generation, thus eliminating the reliance on human labor to collect or draw character images.

\begin{figure*}
  \centering
  \includegraphics[width=0.985\linewidth]{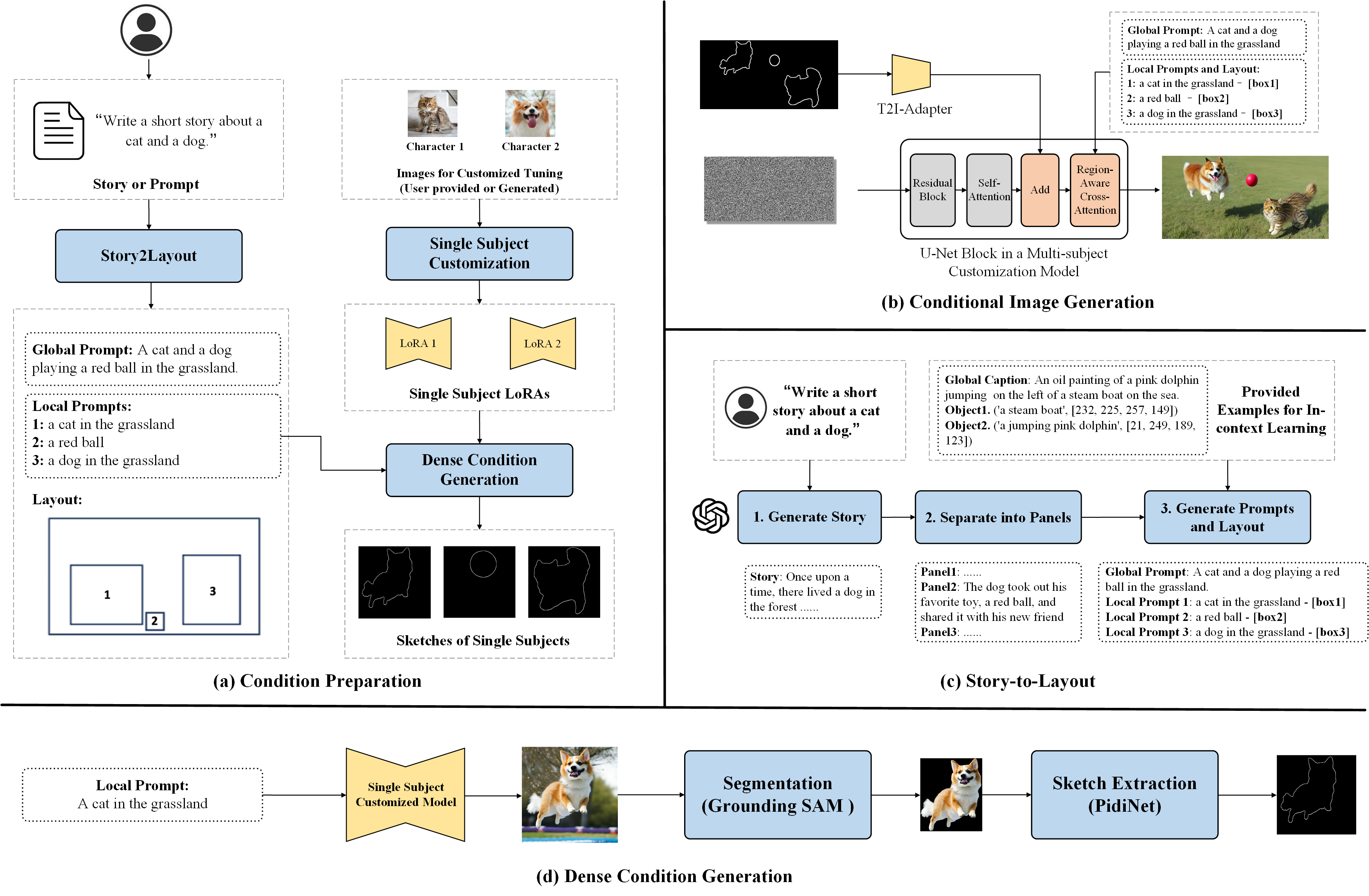}
  \caption{\textbf{The overall pipeline of our proposed method}. The user only needs to provide a short command describing the story and optionally a few images for each character. The pipeline can be roughly divided into (a) the condition preparation stage, where we generate the bounding box layout with corresponding text prompts and the sketch or keypoint dense conditions, and (b) the conditional image generation stage, where we leverage a multi-subject customization model for story images generation, under the guidance of the prepared conditions. 
  The story-to-layout and dense condition generation modules are detailed in (c) and (d), respectively. 
  Specifically, we utilize the LLM for prompt and layout generation in (c) and leverage off-the-shelf perception models to extract dense control signals from object images generated by the single-subject customization model in (d). Both layouts and sketches are easy to understand and manipulate for user interactions.
  }
  \Description{Description}
  \label{fig:pipeline}
\end{figure*}

\section{Our Method}
\label{sec:method}

The goal of our method is to generate diverse storytelling images of high quality and with minimal human effort. Considering the complexity of scenes in storytelling images, our general idea is to combine the comprehension and planning capabilities of LLMs    \cite{openai2023gpt4,anil2023palm} and the generation ability of the large-scale text-to-image models  \cite{rombach2022high}. The pipeline is shown in Fig.~\ref{fig:pipeline}, which can be divided into a condition preparation stage in (a) and a conditional image generation stage in (b).
Specifically, we first utilize LLMs to convert the textual descriptions of stories into layouts of the storytelling images, as detailed in Sec.~\ref{subsec:story2layout}. To improve the quality of generated story images, we propose a simple yet effective method to transform sparse bounding boxes into dense control signals like sketches or keypoints, without introducing manual labor (detailed in Sec.~\ref{subsec:sparse2dense}). Subsequently, we generate story images with a reasonable scene arrangement based on the layout, as detailed in Sec.~\ref{subsec:layout2image}. Finally, we propose a method to eliminate the requirement for users to collect training data for each character, enabling the generation of identity-consistent story images from only texts (detailed in Sec.~\ref{subsec:eliminating}). Since our approach only fine-tunes the pre-trained text-to-image image diffusion model on a few images, we can easily leverage existing models on civiati\footnote{\tt https://civitai.com/} for storytelling in arbitrary characters, scenes, and even styles.

\subsection{Story to Layout Generation}
\label{subsec:story2layout}

\paragraph{Story Pre-processing.}
The user input texts can be either a written story $S$ or a simple description of the story $D$, like ``Write a short story between a bird and a teddy bear”. When only a simple description $D$ is provided as input, we utilize an LLM to generate the specific storylines, {\em i.e.}, $S=\operatorname{LLM}\left(F_{D2S}, D\right)$, as shown in Fig.~\ref{fig:pipeline} (c). Here, $F_{D2S}$ is the instruction that helps the language model to generate the story, {\em e.g.}, ``you are a story writer.'' After obtaining the story $S$, we ask the LLM to segment the story into $K$ panels, each corresponding to a storytelling image, as follows:
\begin{equation}
    \left[P_1, P_2, \ldots, P_K\right]=\operatorname{LLM}\left(F_{S 2 P}, S, K\right),
\end{equation}
where $F_{S2P}$ is the instruction that guides the model to generate panels from the story, and $P_i$ is the textual description of the $i$-th panel. At this point, we have completed the pre-processing of the story.

\paragraph{Layout Generation}

After dividing the story into panel descriptions, we leverage LLMs to extract the scene layout from each panel description, as shown in the following equation:
\begin{equation}
\label{eq:panel2layout}
    \left[\sigma_1, \sigma_2, \ldots, \sigma_K\right]=\operatorname{LLM}\left(F_{P2L},\left[P_1, P_2, \ldots, P_K\right]\right),
\end{equation}
where $F_{P2L}$ is the instruction that guides the model to generate layouts from panel descriptions. Specifically, we provide multiple examples of scene layouts in the instruction to strengthen the LLMs' comprehension and planning ability through in-context learning   \cite{brown2020language}. In this process, we ask the LLM not to use pronouns, such as ``he, she, they, it'', to refer to characters, but instead to specify the name of each subject. In this way, the ambiguity of character references is dramatically reduced. The detailed instructions are shown in Appendix~\ref{subsec:app_details}.

In Eq.~\eqref{eq:panel2layout}, $\sigma_i$ is the scene layout of the $i$-th panel, which consists of a global prompt $p_{i}^{\text{global}}$ and several local prompts with corresponding localized bounding boxes, {\em i.e.}:
\begin{equation}
    \sigma_i=\left\{p_{i}^{\text{global}},\left(p_{i1}^{\text{local}}, b_{i1}\right),\left(p_{i2}^{\text{local}}, b_{i2}\right), \ldots,\left(p_{ik_i}^{\text{local}}, b_{ik_i}\right)\right\},
\end{equation}
where $k_i$ is the number of local prompts in the $i$-th story image. $p_{ij}^{\text{local}}$ and $b_{ij}^{\text{local}}$ are the $j$-th local prompt and bounding box in the $i$-th story image, respectively. While the global prompt describes the global context of the entire story image, the local prompts focus on the details of a single object. 
This design helps us to dramatically improve the quality of image generation by decoupling the complexity of story image generation into multiple simple tasks, as detailed in Sec.~\ref{subsec:sparse2dense} and Sec.~\ref{subsec:layout2image}.

\subsection{Dense Condition Generation}
\label{subsec:sparse2dense}

\paragraph{Motivation.}
Although using sparse bounding boxes as a control signal can improve the generation of subjects and obtain more reasonable scene layouts, we 
find 
that it cannot consistently produce high-quality generation results. There are cases where the images do not exactly match the scene layout or the generated images are of low quality, as detailed in the experiments in Sec.~\ref{subsec:ablations}.

We believe that this is mainly due to the limited information provided by the bounding boxes. The model faces difficulties in generating a large amount of content all at once, with limited guidance. For this reason, we 
propose to 
improve the final story image generation by introducing dense sketch or keypoint guidances. To this end, we devise a \textit{dense condition generation module} based on the layout generated in the previous section, as shown in Fig.~\ref{fig:pipeline}(d).

\paragraph{Subject Generation.}
To transform the sparse bounding box representation of the layout into dense sketch control conditions without introducing human labor, we first generate individual objects in the layout one by one based on the local prompts. 
The process can be represented as: $I_{i j}= 
\operatorname{DM} 
\left(p_{i j}^{l o c a l}\right), j=1,2,...,k_i$, where $I_{ij}$ denotes the $j$-th subject in the $i$-th panel. Thanks to the simplicity of the prompt for single-object generation, the generation process is relatively easy. Thus we are able to obtain high-quality single-object generation results.

\paragraph{Extracting Per-Subject Dense Condition.}

After obtaining the generation results of individual objects, we use the open-vocabulary object detection method, Grouning-DINO   \cite{liu2023grounding}, to localize the object described by the local prompt and obtain the localization box $b_{ij}^{det}$. Afterward, we use SAM   \cite{kirillov2023segment} to obtain the segmentation mask $m_{ij}$ of the object, with $b_{ij}^{det}$ 
being 
the prompt to SAM. Subsequently, following T2I-Adapter, we use PidiNet   \cite{su2021pixel} to obtain the outer edges of the mask, which can be used as the dense sketch for controllable image generation. For the human characters, we can also use HRNet   \cite{wang2020deep} to obtain the human pose keypoints as dense conditions. The control condition corresponding to $I_{ij}$ can be denoted as $C_{ij}$.
It is worth noting that the generated dense control signals are easy to understand and manipulate. Thus, it is easy for the users to 
manually adjust the generated sketches or keypoints to better align with their intentions, if needed.

\paragraph{Composing Dense Conditions}

Lastly, we paste the obtained dense control condition for single objects into their corresponding bounding box regions in the layout to obtain the dense condition for the whole image, denoted as $C_i$. A potential %
issue 
is that the size of the localization box $b_{ij}$ generated by LLM is not exactly the same as the size of the localization box $b_{ij}^{det}$ detected by the Grounding-DINO method 
\cite{liu2023grounding}.
To cope with this, we scale the dense control condition within $b_{ij}^{det}$ to the size of $b_{ij}$ to keep the global layout of the scene unchanged. The process can be written as:
\begin{equation}
    C_i=\operatorname{Compose}\left(l_i,\left[C_{i 1}, C_{i 2}, \ldots, C_{i k_i}\right]\right).
\end{equation}
Note 
that the process of composing dense conditions is fully automatic and does not require any 
manual interaction.

\subsection{Controllable Storytelling Image Generation}
\label{subsec:layout2image}

Large-scale pre-trained text-to-image models, such as Stable Diffusion \cite{rombach2022high}, are capable of generating images from text. However, limited by the language comprehension ability of the text encoder in the model, and the incorrect association between text and image regions in the generation process, the directly generated images often suffer from a series of problems such as object missing, attribution confusion, etc   \cite{chefer2023attendandexcite}. To tackle this, we introduce additional control signals to improve the quality of image generation.

\paragraph{Sparse Layout Control.}

In Sec.~\ref{subsec:story2layout}, we utilized LLMs to obtain the overall layout of the story images. Here, we generate the detailed content of the story images that follow the guidance of the scene layouts. Several existing works have explored generating images using the layout control signal, such as GLIGEN   \cite{li2023gligen}, attention refocus   \cite{phung2023grounded}, BoxDiff   \cite{xie2023boxdiff}, etc. Although all these approaches are applicable, we choose to use the simple and effective region sample approach   \cite{gu2023mix} because it does not introduce any additional model parameters or optimization processes. %
Specifically, in cross-attention, the feature inside the box $b_{ij}$ is replaced by $\operatorname{Attn}(W^Q z_{ij}, W^K E(p_{ij}^{\text{local}}), W^V E(p_{i}^{\text{local}}))$. In this way, we force the image latent feature inside each box to focus on the corresponding local object. Thus we generate images that confirm the layout and also avoid attribute confusion among objects.
The entire process of generating the story image based on the global prompt and sparse bounding box layouts can be written as $\operatorname{DM} \left(p_i^{g l o b a l}; \sigma_i\right)$.

\paragraph{Dense Control.}
To further improve the image quality, we introduce dense conditions generated in Sec.~\ref{subsec:sparse2dense} to guide the image generation process. Specifically, we use the lightweight T2I-Adapter to inject the dense control signals. 
The conditional generation process can be represented as
\begin{equation}
    \operatorname{DM} \left(p_i^{g l o b a l}; \sigma_i,\left\{A, C_i\right\}\right),
\end{equation}
where $C_i$ is the dense condition for the $i$-th story image, $A$ is the T2I-Adapter model for dense control. Unlike TaleCraft    \cite{gong2023talecrafter} which relies on user-input sketches as conditions for every character in each story image, our dense conditions are generated automatically, thus eliminating the tedious process of drawing sketches by hand.

\paragraph{Identity Preservation.}
Identity preservation of the characters plays an important role in achieving visually pleasing story visualization results. We achieve this by %
borrowing the idea of 
Mix-of-Show   \cite{gu2023mix}, as it can preserve the subject identity nicely in a lightweight manner, and is very flexible for multi-concept customization. Specifically, given several images of a subject, a lightweight ED-LoRA    \cite{gu2023mix} weight is fine-tuned for each subject to capture the detailed subject characteristics. Afterward, the gradient fusion    \cite{gu2023mix} is applied to merge multiple ED-LoRAs for individual characters, to guarantee the identity of all characters in the story. The fused LoRA weight %
is 
denoted as $\Delta W$, and the final generation process can be written as:
\begin{equation}
    \operatorname{DM} \left(p_i^{g l o b a l}; \; 
    \sigma_i,\left\{A, C_i\right\}, \Delta W\right).
\end{equation}

\subsection{Eliminating Character-wise Data Collection}
\label{subsec:eliminating}

\begin{figure*}[t]
  \centering
  \includegraphics[width=.73\linewidth]{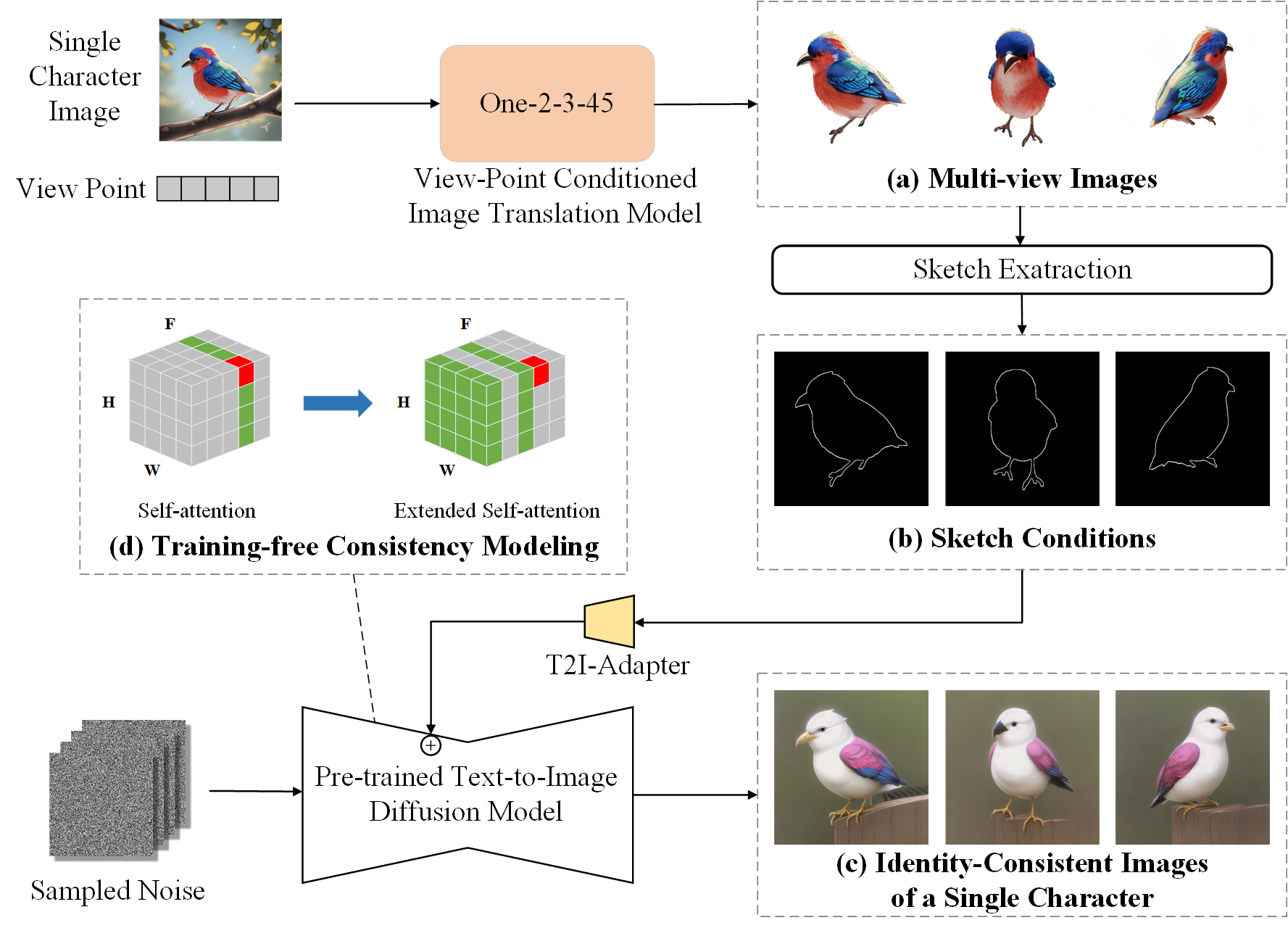}
  \caption{\textbf{Identity-consistent character image generation.}
  To generate multiple identity-consistent images of a single character in (c), we first generate a single character image, then apply a view-point conditioned image translation model to obtain the multi-view images in (a). Afterward, we extract the sketch conditions of those images in (b) and use them as conditions to improve the diversity of the final character image generation. A training-free consistency modeling method is introduced to improve identity consistency in (d). 
  }
  \Description{Description}
  \label{fig:mvdata}
\end{figure*}

\paragraph{The Requirement of Character Data.}

To train a customized model of a character in a story, we need several images of the character for model fine-tuning, which can be written as $\{I_{i}^{s u b}\}, i=1,2,...,n$, where $n$ is the number of images. Existing story visualization methods rely on user-captured images or even datasets to train customized models of characters. 
To eliminate the cumbersome data collection and automate story visualization, we propose a simple and effective method to automatically generate the required training data.
In order to obtain an effective customized model for a single character, the training data needs to satisfy: (1) identity consistency, the structure and texture of the character should be consistent across training images; (2) diversity, the training data should vary, for example in viewpoints, to avoid model overfitting.

\paragraph{Identity Consistency.}

We propose a training-free consistency modeling method to meet the requirement of identity consistency, as shown in Fig.~\ref{fig:mvdata} (d). Specifically, we treat multiple images of a single character as different frames in a video and generate them simultaneously using a pre-trained diffusion model. In this process, the self-attention in the generative model is expanded to other ``video frames"   \cite{wu2022tune,vid2vid-zero} to strengthen the dependencies among images, thus obtaining identity-consistent generation results. 
Concretely, in self-attention, we let the latent features in each frame attend to the features in the first and previous frames to build the dependency. The process can be represented as:
\begin{equation}
    \operatorname{Attn}(W^Q z_{i}, W^K [z_{0}, z_{i-1}], W^V [z_{0}, z_{i-1}]),
\end{equation}
where $z_{i}$ is the latent feature of the current frame, while $z_{0}$ and $z_{i-1}$ are latent features of the first and previous frame, respectively. Here, $[\cdot, \cdot]$ is the concatenation operation.

\paragraph{Diversity.}

Although the above method can ensure the identity consistency of the obtained images, the diversity is not enough for training customized models. For this reason, we inject various conditions in different frames to enhance the diversity of the generated character images. To obtain these diverse yet identity-consistent conditions, we first generate a single image by $I_{i}^{{cond}}=\operatorname{DM} (p_{i}^{{sub}})$, where $p_{i}^{{sub}}$ is the description of the character generated by LLM. Then, we use the pre-trained view-point conditioned image translation model \cite{liu2023zero1to3,liu2023one} to obtain the images of the character from different viewpoints, as shown in Fig. \ref{fig:mvdata} (a). Finally, we extract the sketches or keypoints of these images as the control conditions.

Specifically, for the $i$-th character image, we randomly generate the relative camera rotation $R_{ij} \in \mathbb{R}^{3 \times 3}$ and the relative translation $T_{ij} \in \mathbb{R}^3$ of the desired viewpoint. Then, we use One-2-3-45 to generate the object's images in the desired viewpoints:
\begin{equation}
    I_{ij}^{cond}=f\left(I_{i}^{cond}, R_{ij}, T_{ij}\right), j=1,2, \ldots, n.
\end{equation}
Subsequently, we extract sketches for non-human characters and keypoints for human characters from these images. Finally, we use T2I-Adapter to inject the control guidance into the latent feature of corresponding frames in the generation process.

In addition, in order to further ensure the quality of the generated data, we use CLIP score to filter the generated data, and select the images that are consistent with the text descriptions as the training data for customized generation.

\paragraph{Discussion.}
In this section, we combine the proposed training-free identity-consistency modeling method with the view-point conditioned image translation model to achieve both identity consistency and diversity in character generation. A simpler approach is to directly use the multi-view images from the view-point conditioned image translation model as training data for customization. However, we found that the directly generated results often suffer from distortions or large differences in the color and texture of the images from different viewpoints (see Sec.~\ref{subsec:ablations} for details). For this reason, we need to leverage the above consistency modeling approach to obtain both texture- and structure-consistent images for each character.

\begin{figure*}
  \centering
  \includegraphics[width=0.985\linewidth]{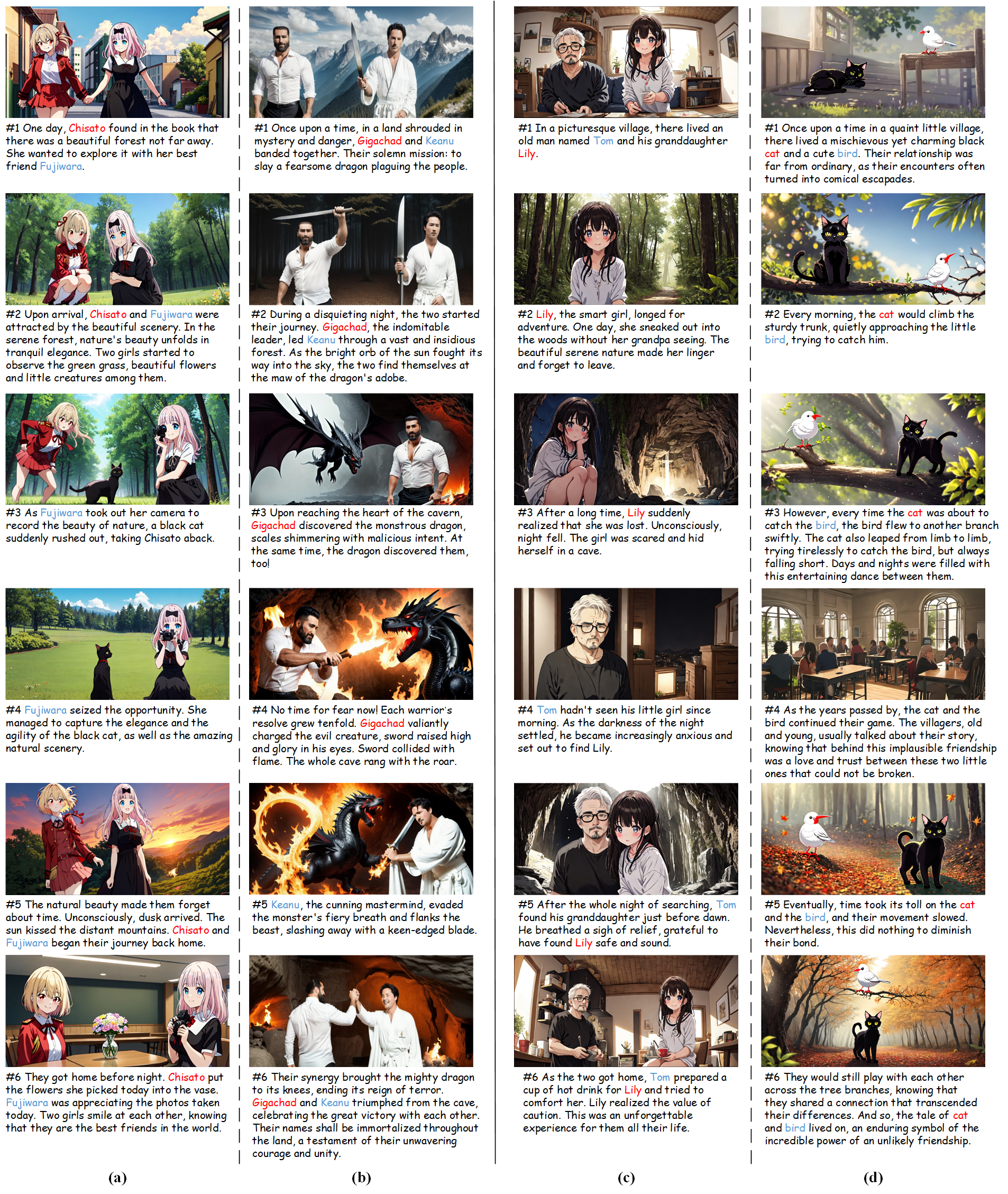}
  \caption{\textbf{A few storytelling results}. Texts below images are the plots of each panel. 
  (a) and (b) are obtained with both user-provided story and character images, while (c) and (d) are obtained with only story text input. The user-provided or generated characters are presented in Appendix~\ref{subsec:app_details}.
  }
  \Description{Description}
  \label{fig:main_results}
\end{figure*}

\begin{figure*}
  \centering
  \includegraphics[width=1\linewidth]{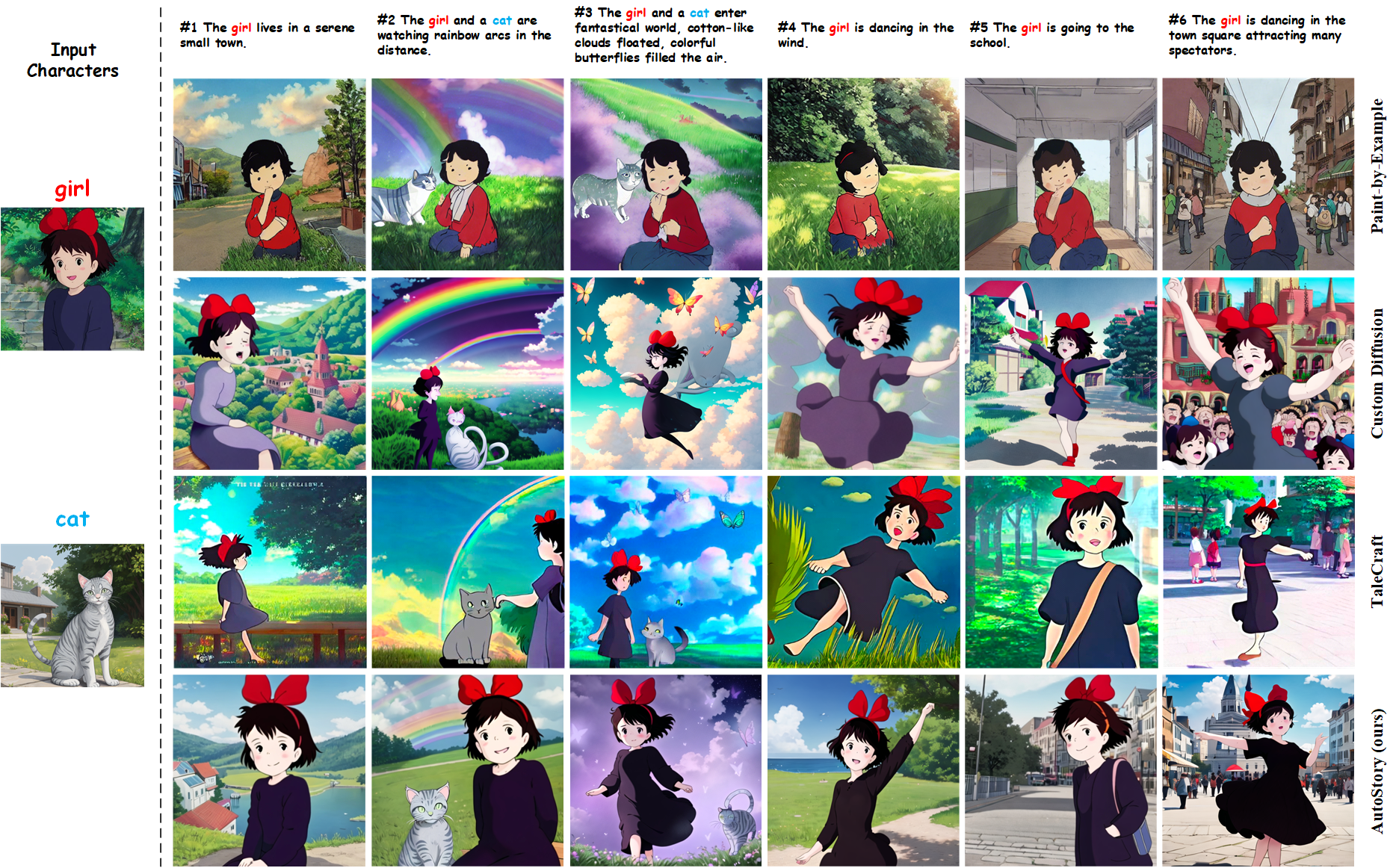}
  \caption{\textbf{Comparison with existing story visualization methods.} 
  The input characters are shown on the left. Note the results of TaleCrafter  \cite{gong2023talecrafter} are directly taken from their paper.
  }
  \Description{Description}
  \label{fig:comparison_cat_girl}
\end{figure*}

\section{Experiments}

\subsection{Implementation Details}
\label{subsec:implementation}

By default, we use GPT-4    \cite{openai2023gpt4} as the LLM for the story to layout generation. The detailed prompts are shown in Appendix~\ref{subsec:app_details}. We use Stable Diffusion   \cite{rombach2022high} for text-to-image generation and leverage existing models on the civitai website
as the base model for customized generation. For dense control, we use T2I-Adapter   \cite{mou2023t2i} keypoint control for human characters, and sketch control for non-human characters. In our \Ours, the only part that requires training is the multi-subject customization process, which takes about 20 minutes for ED-LoRA training and 1 hour for gradient fusion on a single NVIDIA 3090 GPU, while other parts in our pipeline are completely training-free. With the multi-subject customized model prepared, our pipeline can generate plenty of results in minutes.

\subsection{Main Results}
\label{subsec:main_results}

Our AutoStory supports generating stories from user-input text only, or the user can additionally input images to specify the characters in the story. To validate the generality of our approach, we consider story visualization with different characters, scenes, and image styles.
For each story, the text input for the LLM is just one sentence like ``Write a short story about a dog and a cat''. For human characters, we additionally declare their names in the input, {\em e.g.}, ``Write a short story about 2 girls. Their names are Chisato and Fujiwara''. Each character is trained with 5 to 30 images, and the input characters are shown in Appendix~\ref{subsec:app_details}.

\paragraph{With Character Sample Inputs.}

As shown in the first two columns of Fig.~\ref{fig:main_results}, our approach is able to generate high-quality, text-aligned, and identity-consistent story images. Small objects mentioned in the stories are also generated effectively, such as the camera in the third and fourth rows in (a). We attribute this text comprehension and planning capabilities of the LLM, which provides a reasonable image layout without ignoring the key information in the text. 
The features of the characters in each story are highly consistent, including the characters' hairstyles, attire, and facial features. In addition, our approach is able to generate flexible and varied poses for each character, such as the half-squatting position in the third row in (a), and a high-five pose in the last row in (b). This is mainly due to our automatically generated dense control conditions, which guide the diffusion model to obtain fine-grained generation results.

\paragraph{With Only Text Inputs.} 

In the case of text input only, we use the method in Sec.~\ref{subsec:eliminating} to automatically generate training data for each character in the story. The generated character data is shown in Appendix~\ref{subsec:app_details}. As can be seen from the third and fourth columns in Fig.~\ref{fig:main_results}, we are still able to obtain high-quality story visualization results with highly consistent character identities even with only text inputs. The details of the characters in the story images are will-aligned to the text descriptions, {\em e.g.}, the grandfather looks worried when the granddaughter gets lost, in the third and fourth rows in (c). While in the last row, they both look happy when they are reunited back home. The animal characters also show a variety of poses, for example, in (d), the cat presents varying poses of lying down, standing, or walking. This indicates that our method can generate consistent and high-quality story images of characters even without user input of character training images.

\begin{figure}
  \centering
  \includegraphics[width=1\linewidth]{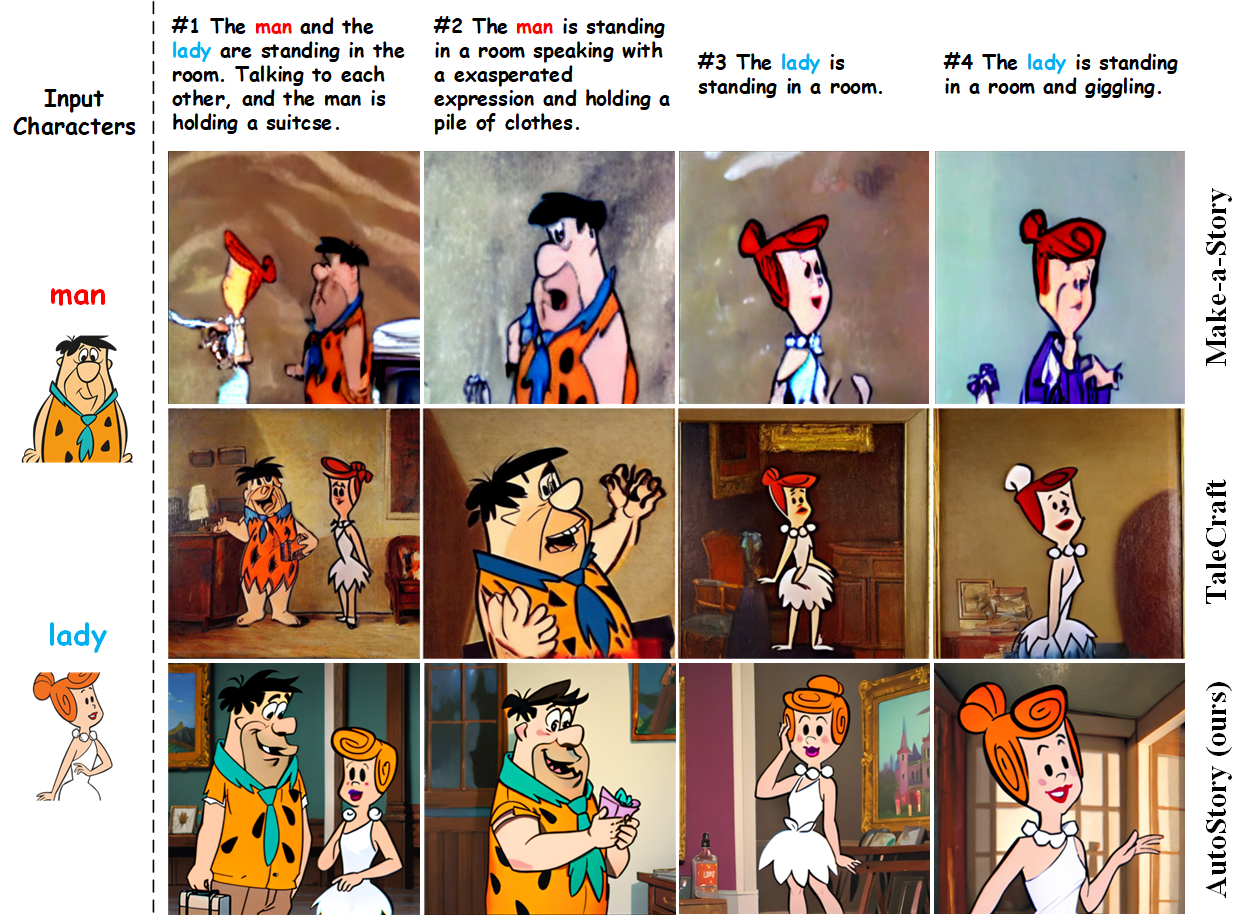}
  \caption{\textbf{Comparison with existing story visualization methods on the FlintstonesSV dataset.} 
  The input characters are shown on the left. Note the results of Make-A-Story \cite{rahman2022make} and TaleCrafter  \cite{gong2023talecrafter} are directly taken from their paper.
  }
  \Description{Description}
  \label{fig:comparison_flintstone}
\end{figure}

\subsection{Comparison with Existing Methods}
\label{subsec:comprisons}

\paragraph{Compared Methods.}
Most previous story visualization methods are tailored for specific characters, scenes, and styles on curated datasets, and cannot be applied to generic story visualization. For this reason, we here mainly compare methods that can generalize, including (1) TaleCraft \cite{gong2023talecrafter}, a very competitive generic story visualization method; (2) Custom Diffusion \cite{kumari2022multi}, a representative multi-concept customization method; (3) paint-by-example \cite{yang2022paint}, which can fill characters into the story image to realize story visualization; (4) Make-A-Story \cite{rahman2022make}, a representative story visualization method in constrained story visualization scenarios, which is compared in the qualitative experiments.
Since all existing methods rely on the user input character images for training, here we consider the same setting for a fair comparison.

\paragraph{Qulitative Comparison.}

In order to make a head-to-head comparison with the existing story visualization methods, we adopt the stories in TaleCraft and Make-A-Story, as shown in Fig.~\ref{fig:comparison_cat_girl} and Fig.~\ref{fig:comparison_flintstone}. It should be noted that since the character training images in TaleCraft are not available, we collected training images for each character in the story. Therefore, the input character images of our approach are slightly different from those used by TaleCraft. As shown in Fig.~\ref{fig:comparison_cat_girl}, Paint-by-example struggles to preserve the identities of characters. The girls in the generated images differ significantly from the user-provided image of the girl. Although Custom Diffusion performs slightly better in identity preservation, it sometimes generates images with obvious artifacts, such as the distorted cat in the second and third images. TaleCraft achieves better image quality but still suffers from certain artifacts, {\em e.g.}, the cat in the third image is distorted and one of the girl’s legs in the fourth image is missing. In contrast, our method is able to achieve superior performance in terms of identity preservation, text alignment, and generation quality.

Similarly, In Fig.~\ref{fig:comparison_flintstone}, it can be seen that Make-A-Story generates story images in low quality, which is mainly due to the fact that it’s tailored for the FlintstonesSV \cite{maharana2021vlcstorygan} dataset, and thus inherently limited by generation capacity. TaleCraft shows significant improvement in generation quality, but it has limited alignment to text, {\em e.g.}, the missing suitcase in the first image, which we assume is due to the limited layout generation capacity of the discrete diffusion model for layout generation. In contrast, our method is able to text-aligned results, thanks to the LLM's strong text comprehension and layout planning capabilities. Interestingly, there are significant differences in image style between our AutoStory and TaleCraft. We hypothesize that this is mainly caused by the difference in character data for training.

\paragraph{Quantitative Comparison.}

\input{tables/table_metric}

Following the literature \cite{gong2023talecrafter}, we consider two metrics to evaluate the generated results: (1) text-to-image similarity, which is measured by the cosine similarity between the embeddings of texts and images in the CLIP feature space; (2) image-to-image similarity, which is measured by the cosine similarity between the average embedding of character images for training and the embedding of generated story images in CLIP image space. We conduct experiments on 10 stories with a total of 71 prompts and corresponding images. The results are shown in Table~\ref{tab:metrics}. It can be seen that our \Ours outperforms existing methods by a notable margin in both text-to-image similarity and image-to-image similarity, which demonstrates the superiority of our method.

\paragraph{User Study.}
\input{tables/table_user}

We conduct user studies on 10 stories, with an average of 7 prompts per story. During the study, 32 participants are asked to rate the story visualization results on three dimensions: (1) the alignment between the text and the images; (2) the identity-preservation of the characters in the images; and (3) the quality of the generated images. We asked users to score each set of story images on a Likert scale of 1-5. The results for each method are shown in Table \ref{tab:user-study}. It can be seen that our \Ours outperforms competing methods by a large margin in all three metrics, which indicates that our method is more favored by users.

\begin{figure*}
  \centering
  \includegraphics[width=1\linewidth]{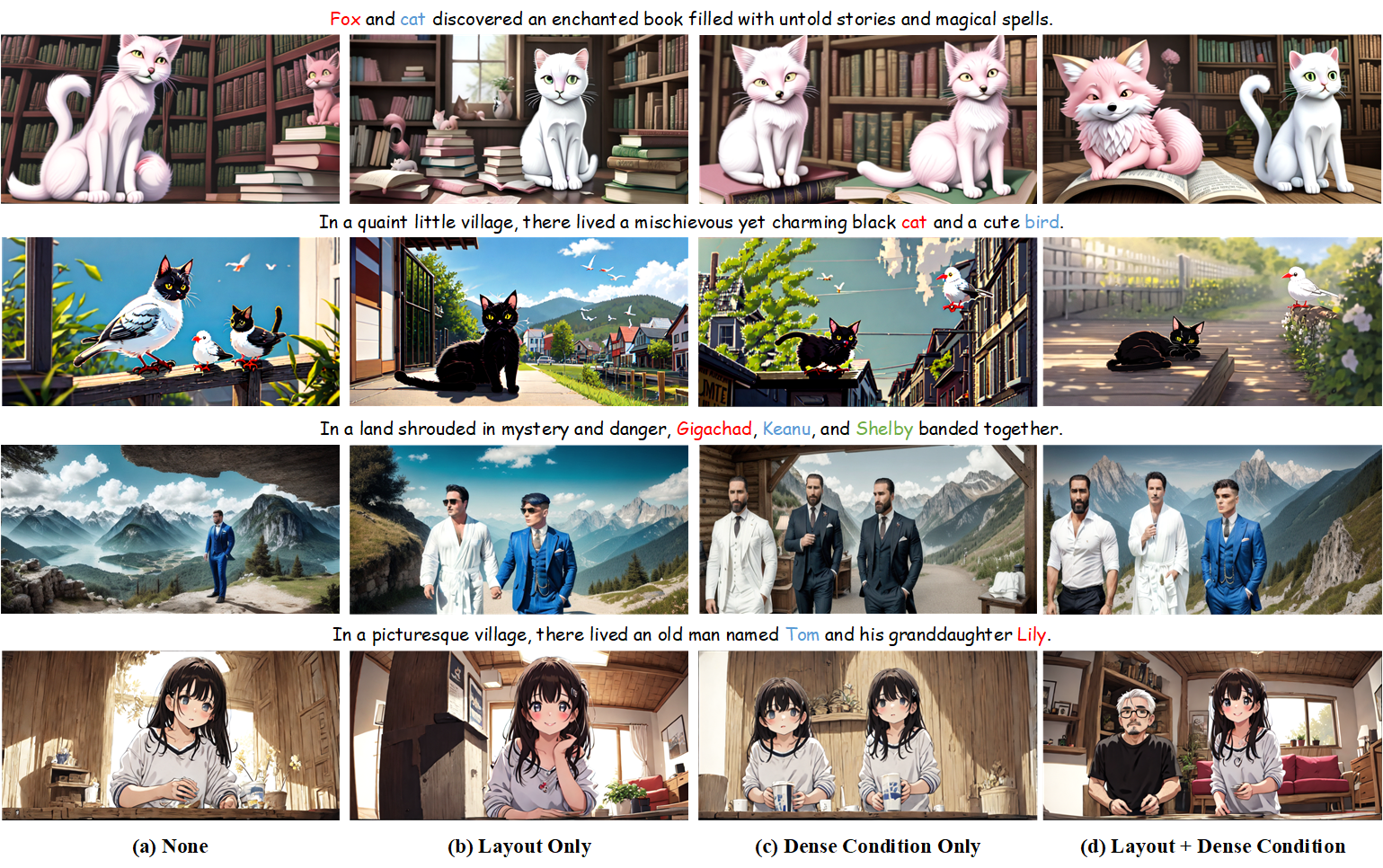}
  \caption{
  \textbf{Ablations on different control strategies.} The first two rows use sketches as the dense condition, while the last two rows leverage keypoints as the dense condition.
  }
  \Description{Description}
  \label{fig:ablation_cond}
\end{figure*}

\begin{figure}
  \centering
  \includegraphics[width=1\linewidth]{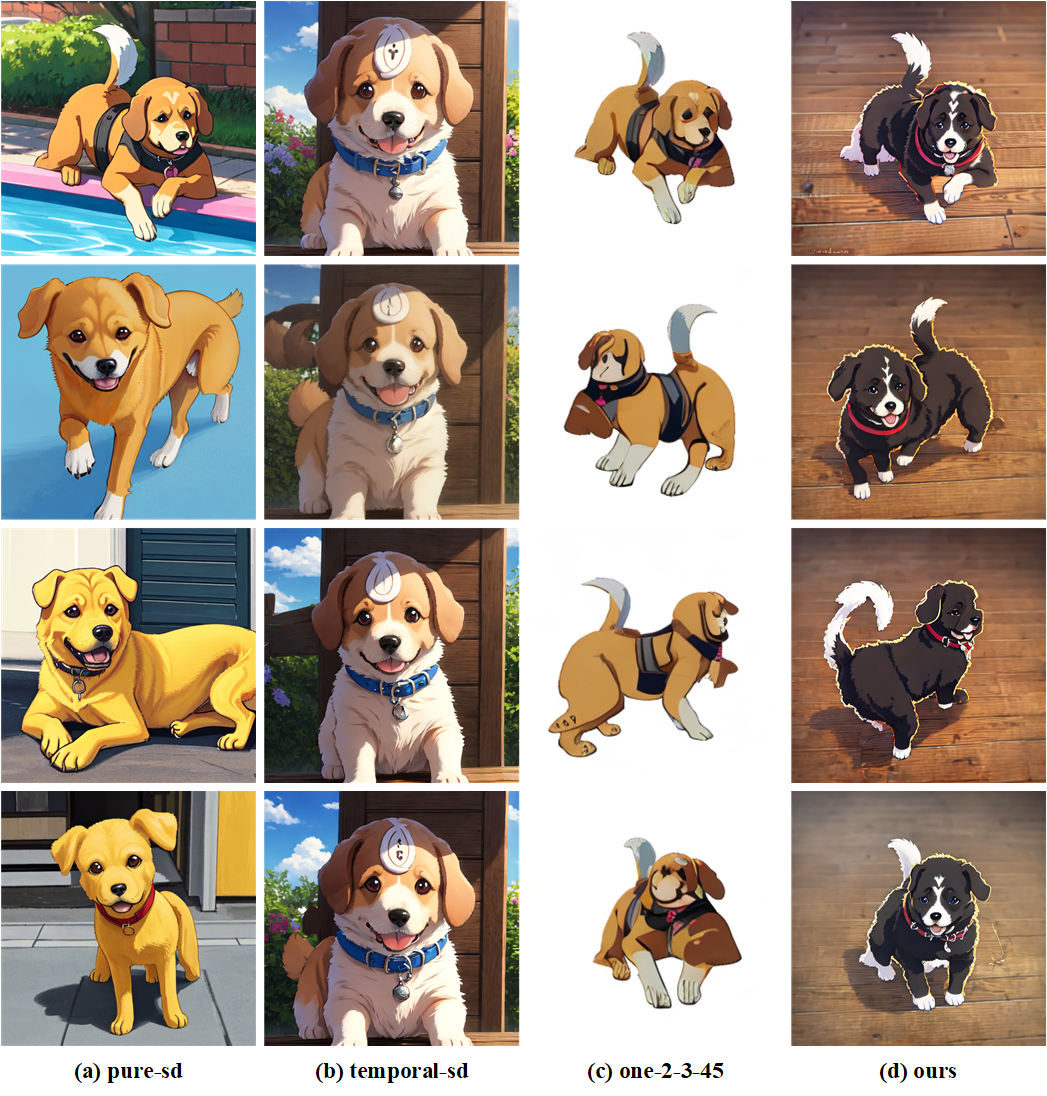}
  \caption{\textbf{Ablations on character data generation.} 
  (a) pure-sd uses the original Stable Diffusion for data generation. (b) temporal-sd generates multiple characters images simultaneously with the extended self-attention in Sec.~\ref{subsec:eliminating}. (c) one-2-3-45 generates character images of varying viewpoints from a single character image. (d) ours combines both extended self-attention and One-2-3-45 for character image generation. 
  }
  \Description{Description}
  \label{fig:ablation_data}
\end{figure}

\subsection{Ablation Studies}
\label{subsec:ablations}

\paragraph{Ablations on Control Signals.}

We evaluate the necessity of both layout control and dense condition control in this section. The layout control refers to the bounding boxes indicating object locations and the corresponding local prompts, while the dense condition control refers to the composed condition, such as sketches and keypoints. The results are shown in Fig. \ref{fig:ablation_cond}, with the first two rows using sketches and the last two rows using keypoints as the dense condition. 
We have the following observations. 
Firstly, when no control conditions are used, the model generates images with missing objects and blends the properties of different objects, as shown in Fig.~\ref{fig:ablation_cond} (a). For example, only one character is generated in the third line, while the other two characters in the text are ignored. In the second line, there is a conflict between the attributes of a cat and a bird, and the generated animal has the head of a cat and the wings of a bird. This is mainly due to the fact that the generative model can not well-capture the textual input to generate images that have proper layouts and differentiate the attributes of the varying entities.
Secondly, with the addition of the layout control, the concept conflict is significantly alleviated, mainly because the layout control helps to associate specific regions in the image with the corresponding local prompts. However, the problem of missing subjects in the images still exists, for example, only two characters are generated in the third row, while the character Gigachad is ignored in Fig.~\ref{fig:ablation_cond} (b). We suspect that this is due to the limited influence of the layout control on the feature updating in the model.
Thirdly, in the case of only adding the dense control condition, the model is able to effectively generate all the entities mentioned in the text without omitting them, mainly because the dense control condition provides sufficient guidance to the model. However, the conceptual conflicts among the characters persist, for example, the attributes of the man in the fourth line are dominated by the attributes of the girl. This is mainly due to the fact that the character regions in the image are incorrectly and strongly associated with the other characters in the text.
Lastly, our approach combines layout and dense conditional control can avoid object omissions and conceptual conflicts among characters, resulting in high-quality story images. We attribute this to the proper layout generated by the LLM and the effective conditioning paradigm during image generation.

\paragraph{Ablations on Designs in Multi-view Character Generation.}
To support the generation of story images from text inputs only, we propose an identity-consistent image generation approach to eliminate character-wise data collection, as detailed in Sec.~\ref{subsec:eliminating}. Here we ablate the design in this module and consider the following baseline approaches for comparison: (1) the pure-sd variant, which generates multiple character images directly using the Stable Diffusion model, without any additional operations. (2) the One-2-3-45 variant, which combines Stable Diffusion and One-2-3-45 for identity-consistent character image generation. Specifically, a single character image is first generated using Stable Diffusion, and then multi-view character images are obtained by applying One-2-3-45 to the single generated image. (3) the temporal-sd variant, which treats multiple character images as a video and leverages the extended self-attention in Sec.~\ref{subsec:eliminating} for training-free consistency modeling. 
Firstly, pure-sd fails to obtain identity-consistent images as training data for a single character. As shown in the first column in Fig.~\ref{fig:ablation_data}, the color and the body shape of dogs in different images vary significantly.
Secondly, the identities of the dogs in the images obtained using temporal-sd are consistent, as shown in the second column. This is because after adding extended self-attention, the latent features of several images can interact with each other, which substantially improves the consistency among images. However, the dogs in these images are all displayed in a positive smiling posture, indicating the lack of diversity.
Thirdly, the images obtained using One-2-3-45 show strong diversity, but suffer from certain artifacts, such as the deformation of the dog's head, as shown in the third column. This is mainly because One-2-3-45 can not guarantee the consistency of the generated multi-view images.
Lastly, our method is able to enhance diversity while ensuring the identity consistency of the generated character images. This is mainly due to the fact that we utilize the sketch of the images obtained by One-2-3-45 to guide the model for generating diverse character data, while using extended self-attention to ensure the consistency among images. In addition, the image priors cherished by Stable Diffusion can substantially mitigate the negative impact caused by the imperfect sketches obtained from images generated by One-2-3-45. As can be seen, the dogs generated by our method are free from distortions.

\section{Conclusion}

The main focus of our \Ours is to create diverse story visualizations that meet specific user requirements with minimal human effort. By combining the capabilities of the LLMs and diffusion models, we managed to obtain text-aligned, identity-consistent, and high-quality story images. Furthermore, with our well-designed story visualization pipeline and the proposed character data generation module, our approach streamlines the generation process and reduces the burden on the user, effectively eliminating the need for users to perform labor-intensive data collection. Sufficient experiments demonstrate that our method outperforms existing approaches in terms of the quality of the generated stories and the preservation of the subject characteristics. Moreover, our superior results are achieved without requiring time-consuming and computationally expensive large-scale training, making it easy to generalize to varying characters, scenes, and styles. In future work, we plan to accelerate the multi-concept customization process and make our \Ours run in real-time.

\bibliographystyle{ACM-Reference-Format}
\bibliography{refs}

\newpage
\appendix
\section{Appendix}

\subsection{More Implementation Details}
\label{subsec:app_details}

\begin{figure*}
    \centering
    \includegraphics[width=0.9\linewidth]{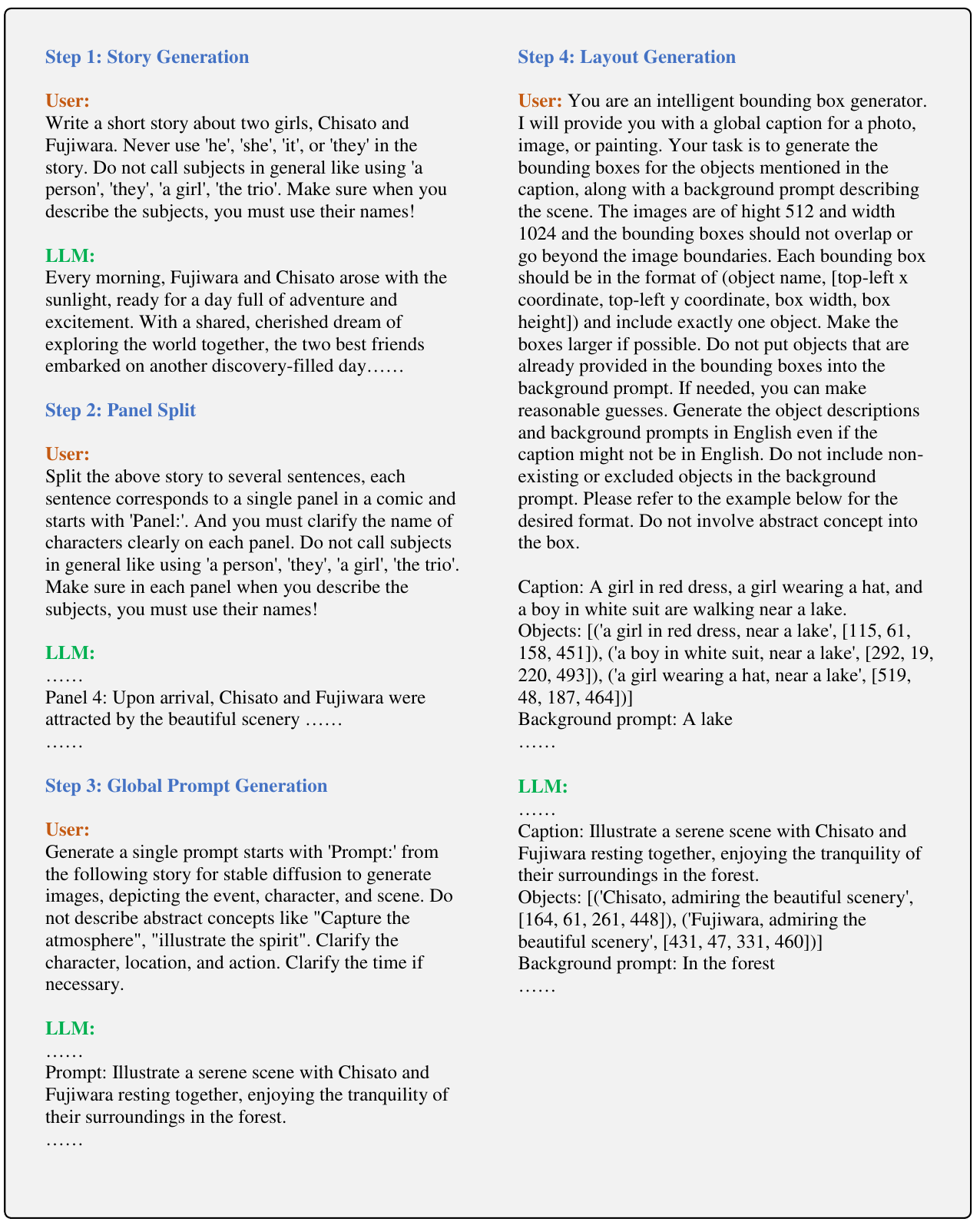}
    \caption{\textbf{Prompts for story and layout generation.} The users only need to provide the story requirements such as ``write a short story about two girls, Chisato and Fujiwara''.
    }
    \label{fig:app_prompts}
\end{figure*}

\paragraph{Detailed Prompts for the LLM.}
As described in Sec.~\ref{sec:method} in the main text, we utilize LLMs to accomplish the story and layout generation. Specifically, we leverage the LLM for (1) generating the story, (2) dividing the story into panels, and (3) generating prompts and layout from the panels.
In implementation, we further split the third step into two sub-steps, where we first convert the text of each panel into prompts suitable for generating the image, and then parse the prompts into the layout and local prompts. The detailed prompts and sampled LLM outputs are shown in Fig.~\ref{fig:app_prompts}.

\begin{figure}
  \centering
  \includegraphics[width=1\linewidth]{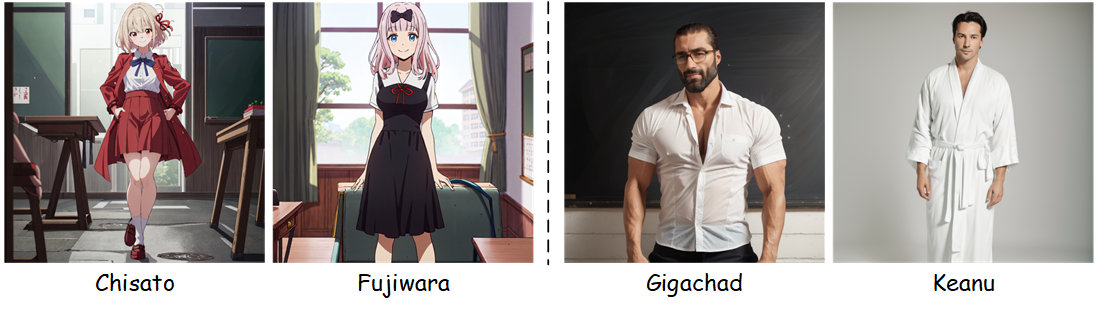}
  \caption{\textbf{User input characters.} }
  \Description{Description}
  \label{fig:app_char_input}
\end{figure}

\begin{figure}
  \centering
  \includegraphics[width=1\linewidth]{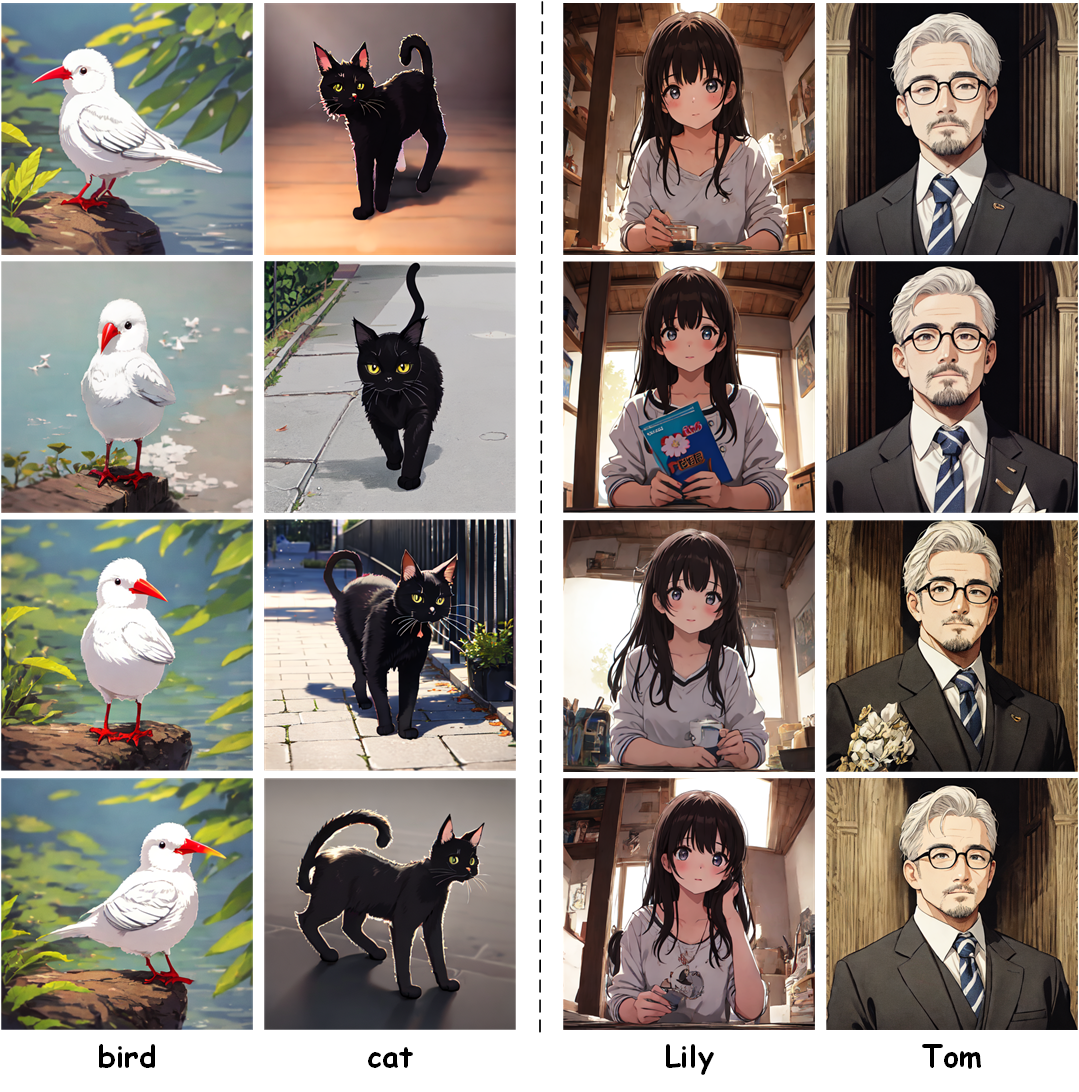}
  \caption{\textbf{Character images generated by our \Ours.} }
  \Description{Description}
  \label{fig:app_char_auto}
\end{figure}

\paragraph{More Details on the Main Results.}
Fig.~\ref{fig:main_results} shows the story image generation results of our method with varying characters, storylines, and image styles. Here, we present the character images used to train the customized model for each story. 
The story visualization results in the left two columns in Fig.~\ref{fig:main_results} are obtained with the user-supplied character images. The corresponding characters are shown in Fig.~\ref{fig:app_char_input}. 
Differently, the story visualization results in the right two columns are obtained with only the story texts as inputs, and the characters are automatically generated by our method. The generated images for each character are shown in Fig.~\ref{fig:app_char_auto}. 
It can be seen that the animal and human characters generated by our method are of high quality and consistent identities. The images of a single animal character show high diversity, with the orientation of the bird and the cat changing constantly from left to right. 
The human characters, however, are slightly less diverse, with a lower degree of variance in facial orientation. We believe that this is mainly due to the fact that the diffusion model is trained primarily on humans with frontal faces, making it difficult to generate side-facing images. Nonetheless, the character image data generated by our method can be effectively used for training customization models in story visualization, without introducing overfitting.
It is worth mentioning that even though the character Tom in our generated data wears suits, we can generate images with Tom wearing a T-shirt after we specify that the character wears a T-shirt in the local prompt, as shown in the story visualization in Fig.~\ref{fig:main_results} (d). Moreover, the characteristics of Tom are well-maintained, such as the shape of his face and the white hair. 
This indicates that the customized model trained with our generated data learns the character's identity without overfitting.

\subsection{Intermediate Results Visualization}
\label{subsec:app_middle_results}

\begin{figure}
  \centering
  \includegraphics[width=1\linewidth]{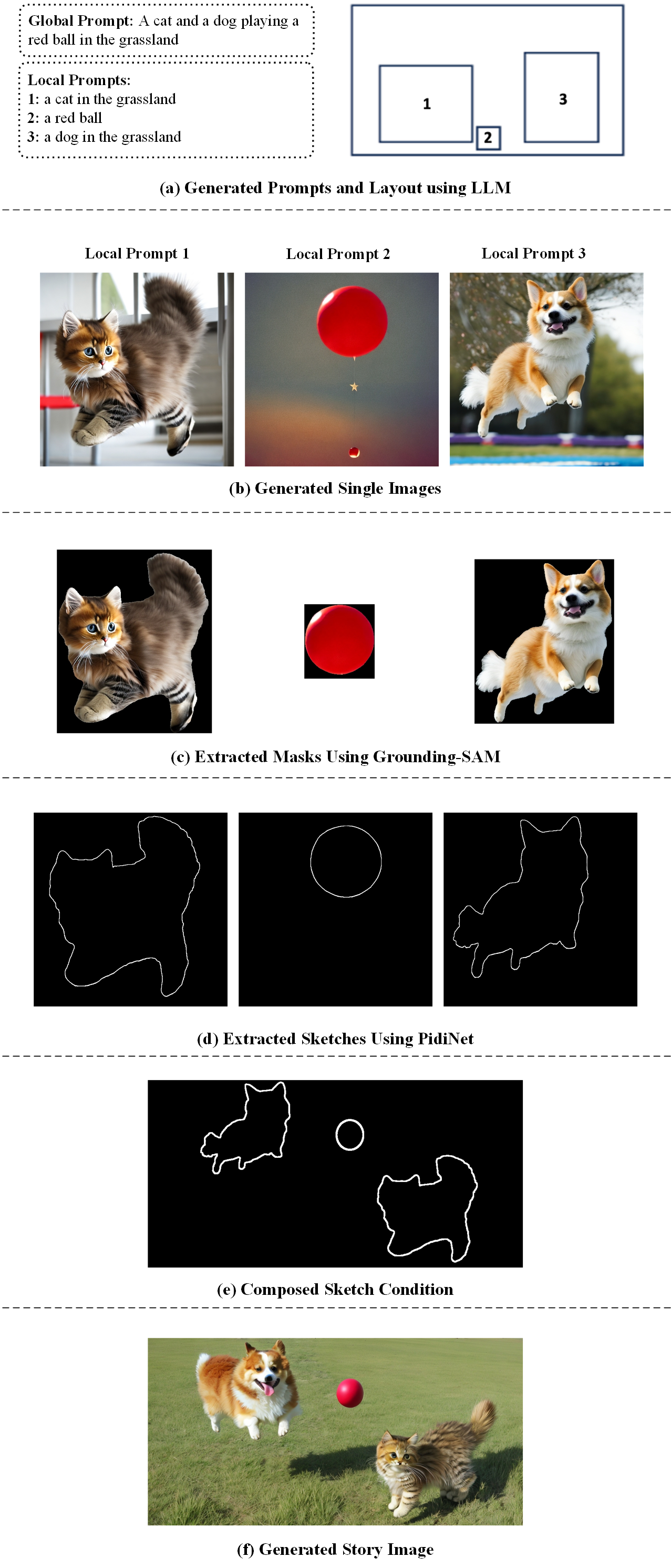}
  \caption{Visualization of intermediate results for generating a single story image. We use sketch conditions for non-human characters and subjects.}
  \Description{Description}
  \label{fig:app_pro_sketch}
\end{figure}

\begin{figure}
  \centering
  \includegraphics[width=1\linewidth]{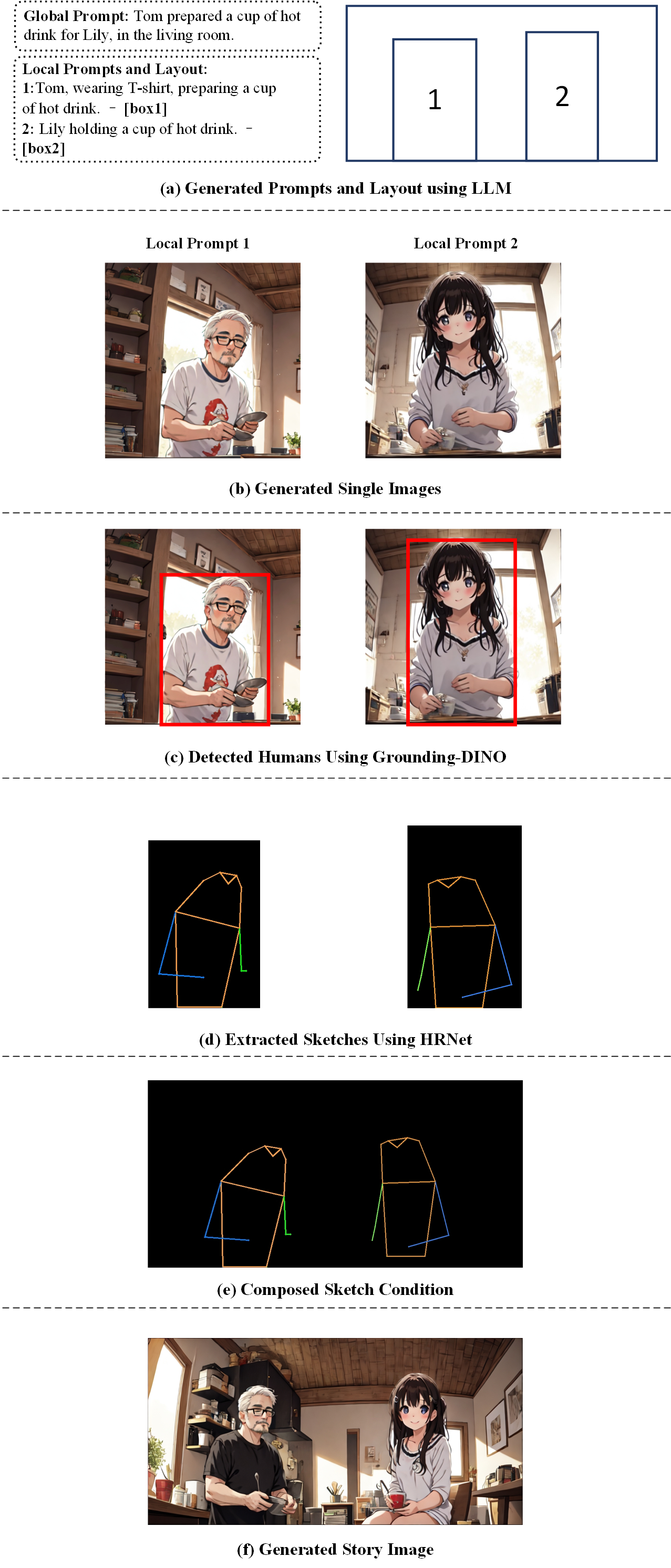}
  \caption{Visualization of intermediate results for generating a single story image. We use keypoint conditions for human characters.}
  \Description{Description}
  \label{fig:app_pro_pose}
\end{figure}

To better understand our approach, in this section, we visualize the intermediate process of generating a single story image, as shown in Fig.~\ref{fig:app_pro_sketch} and Fig.~\ref{fig:app_pro_pose}. We first generate single-character images based on the Local prompts generated by LLM, as shown in (a) and (b). The perception models, including Grounding-SAM, PidiNet, and HRNet, are then utilized to obtain the keypoints of human characters, or sketches of non-human characters, as shown in (c) and (d). Subsequently, the LLM-generated layout is utilized to compose the keypoints or sketches of individual subjects into a dense condition for generating story images, as shown in (d). Finally based on the dense conditions, prompts, and layout, we generate the story image as shown in (f).

\subsection{More Story Visualization Results}
\label{subsec:app_more_demos}

\begin{figure*}[b]
  \centering
  \includegraphics[width=0.9\linewidth]{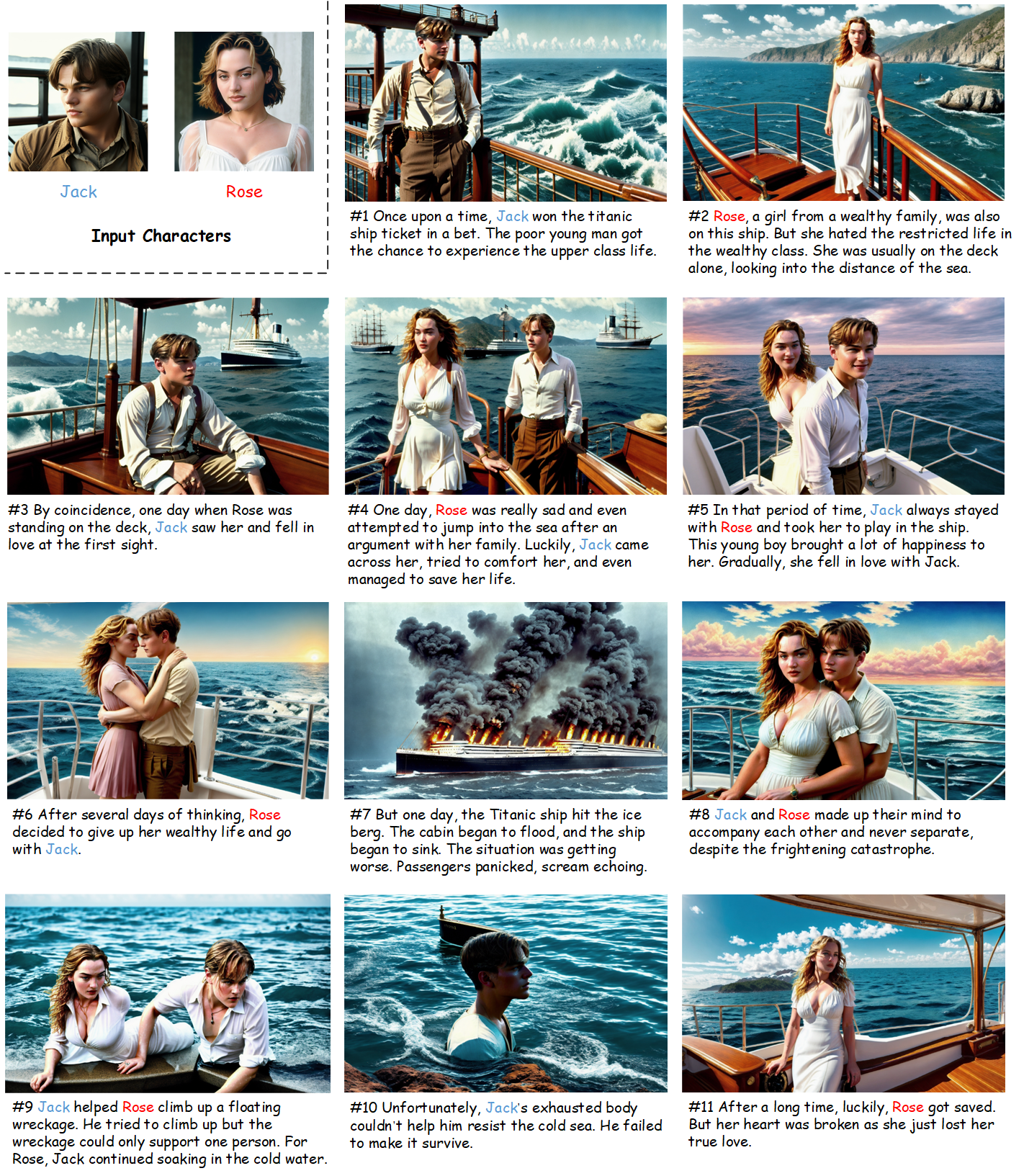}
  \caption{\textbf{More story visualization results.} 
  Note that the texts here are story plots, not the prompts to the diffusion model.}
  \label{fig:app_titanic}
\end{figure*}

\begin{figure*}[b]
  \centering
  \includegraphics[width=\linewidth]{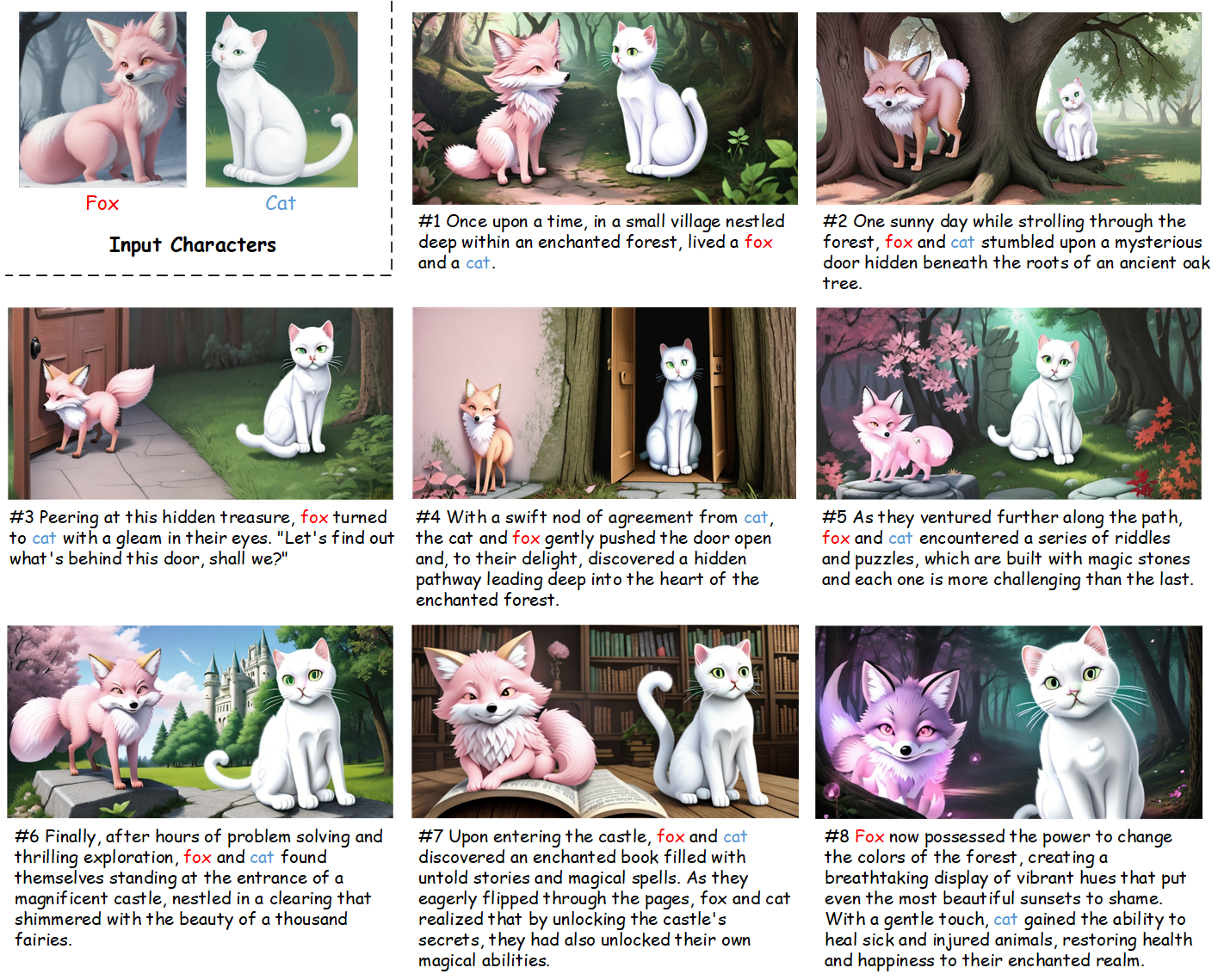}
  \caption{\textbf{More story visualization results.} 
  Note that the texts here are story plots, not the prompts to the diffusion model.}
  \label{fig:app_car_fox}
\end{figure*}

In Fig.~\ref{fig:app_titanic} and Fig.~\ref{fig:app_car_fox}, we showcase more story visualization results of our method. As can be seen, our \Ours can produce high-quality, text-aligned, and identity-consistent story images, even when generating long stories.

\end{document}

%% file: tables/table_metric.tex
\begin{table}[]
\scalebox{1}{
\begin{tabular}{lccc}
\hline
{Method} & Custom-Diffusion & Paint-by-Example & Ours \\ \hline
text-image sim.   & 0.7332                 & 0.7172                 & \textbf{0.7721}     \\
image-image sim.  & 0.6402                 &  0.6214                &  \textbf{0.6748}    \\ \hline
\end{tabular}
}
\caption{\textbf{Quantitative comparisons}. Both text-to-image and image-to-image similarity are computed in the CLIP feature space.}
\label{tab:metrics}
\end{table}

%% file: tables/table_user.tex
\begin{table}[]
\scalebox{1}{
\begin{tabular}{lccc}
\hline
{Method} & Custom-Diffusion & Paint-by-Example & Ours \\ \hline
Correspondence   & 2.19                 & 2.17                 & \textbf{4.31}     \\
Coherence        & 2.64                 & 2.53                 & \textbf{4.16}     \\
Quality          & 2.65                 & 2.35                 & \textbf{4.08}     \\ \hline
\end{tabular}
}
\caption{\textbf{User study results}. Users are asked to rate the results on a Likert scale of 1 to 5 according to text-to-image alignment (Correspondence), identity preservation (Coherence), and image quality (Quality).}
\label{tab:user-study}
\end{table}